\documentclass[lettersize,journal]{IEEEtran}

\usepackage[table,xcdraw]{xcolor}

\usepackage{kotex}
\usepackage{graphicx}
\usepackage{tabularx}
\usepackage{amsmath}
\usepackage{mathtools}
\usepackage{graphicx}
\usepackage{mathtools}
\usepackage{multirow}
\usepackage{hhline}
\usepackage{booktabs}
\usepackage[]{algorithm2e}
\usepackage{mathabx}
\usepackage{eucal}
\usepackage{txfonts}
\usepackage{color,soul}
\usepackage{array}
\usepackage{tikz}
\usepackage{comment}
\usepackage{amssymb} 
\usepackage{color}
\usepackage{cite}
\usepackage{arydshln}
\usepackage{subcaption}

\newcommand{\Tref}[1]{Table~\ref{#1}}

\newcommand{\Fref}[1]{Fig.~\ref{#1}}

\title{
Scene-Agnostic Traversability Labeling and Estimation via a Multimodal Self-supervised Framework
}

\author{Zipeng Fang, Yanbo Wang, Lei Zhao and Weidong Chen}

\begin{document}

\maketitle
\pagestyle{empty}  
\thispagestyle{empty} 

\begin{abstract}
Traversability estimation is critical for enabling robots to navigate across diverse terrains and environments. While recent self-supervised learning methods achieve promising results, they often fail to capture the characteristics of non-traversable regions. Moreover, most prior works concentrate on a single modality, overlooking the complementary strengths offered by integrating heterogeneous sensory modalities for more robust traversability estimation.
To address these limitations, we propose a multimodal self-supervised framework for traversability labeling and estimation. First, our annotation pipeline integrates footprint, LiDAR, and camera data as prompts for a vision foundation model, generating traversability labels that account for both semantic and geometric cues. Then, leveraging these labels, we train a dual-stream network that jointly learns from different modalities in a decoupled manner, enhancing its capacity to recognize diverse traversability patterns. In addition, we incorporate sparse LiDAR-based supervision to mitigate the noise introduced by pseudo labels.
Finally, extensive experiments conducted across urban, off-road, and campus environments demonstrate the effectiveness of our approach. The proposed automatic labeling method consistently achieves around 88\% IoU across diverse datasets. Compared to existing self-supervised state-of-the-art methods, our multimodal traversability estimation network yields consistently higher IoU, improving by 1.6–3.5\% on all evaluated datasets.
\end{abstract}


\section{INTRODUCTION}
\IEEEPARstart{T}{raversability} estimation is a fundamental aspect of robotic perception, aiming to determine which regions of the environment are safe for robots to navigate~\cite{papadakis2013terrain,borges2022survey}. Non-traversable regions generally arise from two major factors. The first involves geometric hazards, such as steep steps or abrupt elevation changes that may cause the robot to tip over. The second relates to semantic constraints, where socially or contextually undesirable behaviors should be avoided. For example, a car should not drive onto pedestrian crosswalks, and a robot should refrain from rolling over grass or flower beds.

Early research has made significant progress in both rule-based and segmentation-based traversability estimation. Rule-based approaches~\cite{Wermelinger_Navigation,Fan2021STEPST,shan2018bayesian} typically extract terrain features such as step, slope, and roughness and use weighted combination schemes to estimate traversability scores. However, these methods are vulnerable to noise and exhibit limited scalability to robots with diverse capabilities.
Alternatively, segmentation-based approaches~\cite{guan2022ga,shaban2022semantic,wang2024sfpnet} employ deep learning models to classify terrain types and assign predefined cost values to each semantic class. Such methods require large-scale labeled datasets and generalize poorly to previously unseen terrains. Moreover, it is inherently difficult to define precise and meaningful cost values for every semantic class.

Recently, self-supervised learning~\cite{zurn2020self,mattamala2024wild,seo2023learning} has emerged as a promising paradigm for traversability estimation due to its strong scalability. These methods typically project the robot's future trajectory onto camera images to generate traversability labels automatically. However, these approaches face intrinsic limitations. On the one hand, time constraints make it infeasible for the robot to cover the entire set of traversable areas. On the other hand, to ensure safety, the robot cannot physically traverse dangerous or unknown regions, making it impossible to collect negative samples from obstacles. Some recent studies have addressed these limitations through contrastive or unsupervised representation learning, while others have leveraged powerful segmentation models, such as the Segment Anything Model (SAM), to expand label coverage~\cite{jung2024v,ma2024imost,kim2024learning}. Nonetheless, most of these methods fail to incorporate explicit negative supervision, and many rely solely on a single modality, neglecting the different contributions of semantic and geometric cues to traversability understanding.


\begin{figure}[t!]
\centering
\begin{tabular}{*{3}{c@{\hspace{3px}}}}
\begin{subfigure}[b]{0.30\linewidth}
    \includegraphics[width=\linewidth]{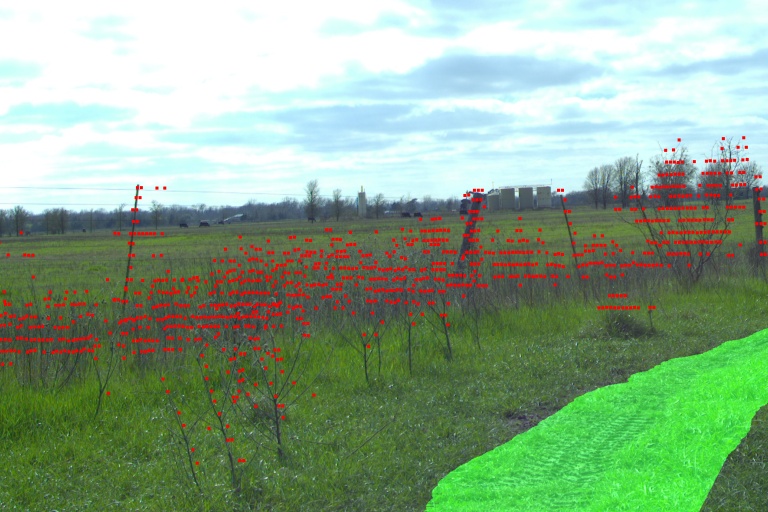}
\end{subfigure} &
\begin{subfigure}[b]{0.30\linewidth}
    \includegraphics[width=\linewidth]{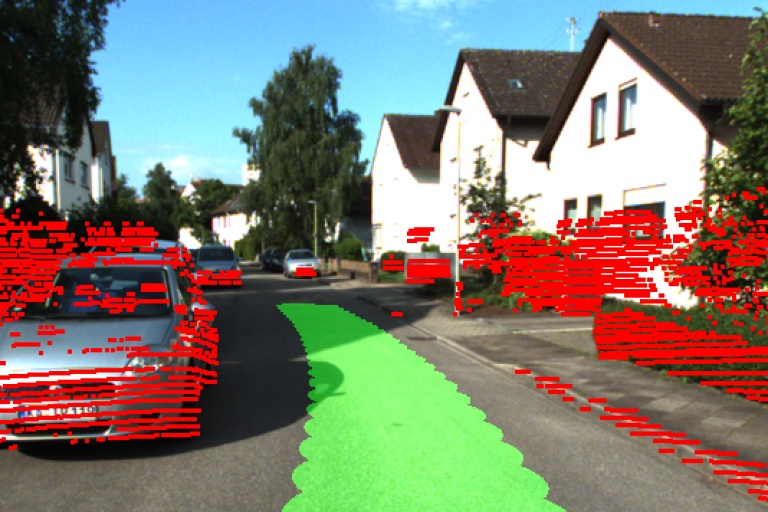}
\end{subfigure} &
\begin{subfigure}[b]{0.30\linewidth}
    \includegraphics[width=\linewidth]{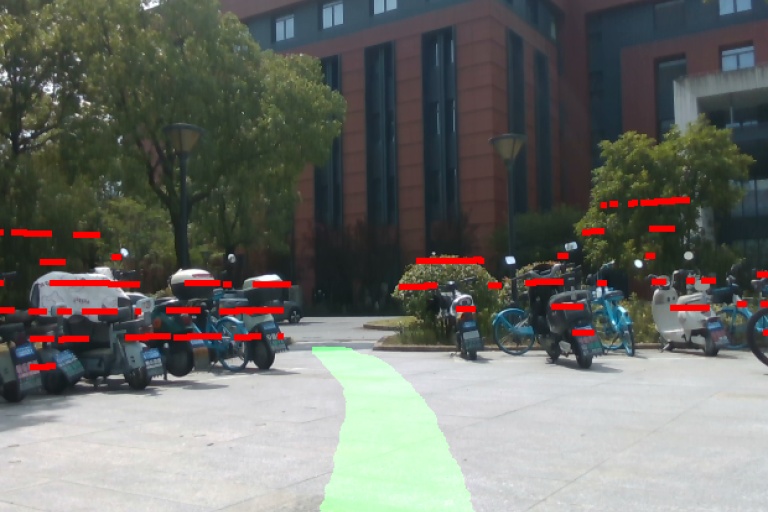}
\end{subfigure} \\
\begin{subfigure}[b]{0.30\linewidth}
    \includegraphics[width=\linewidth]{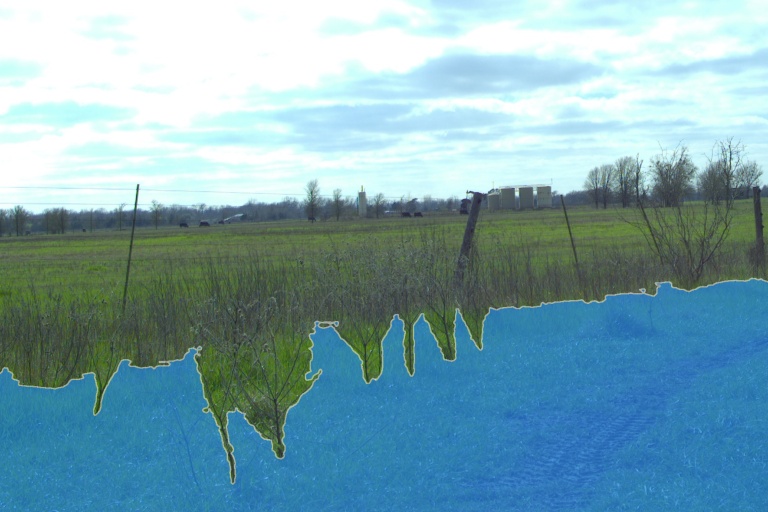}
\end{subfigure} &
\begin{subfigure}[b]{0.30\linewidth}
    \includegraphics[width=\linewidth]{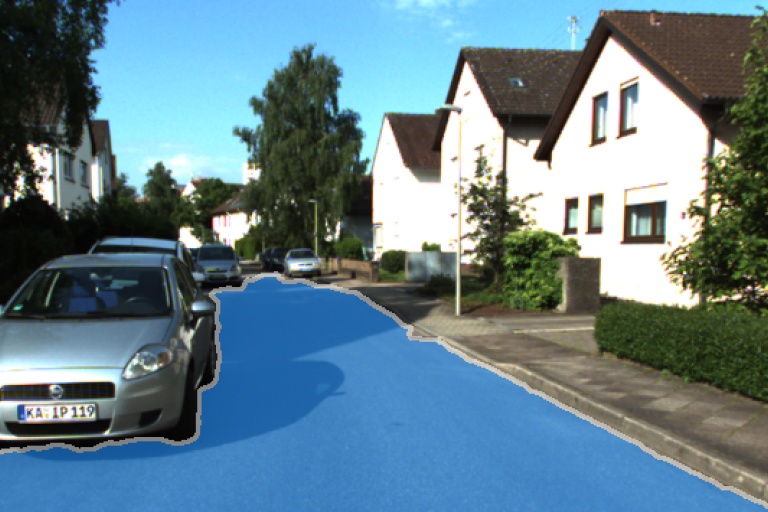}
\end{subfigure} &
\begin{subfigure}[b]{0.30\linewidth}
    \includegraphics[width=\linewidth]{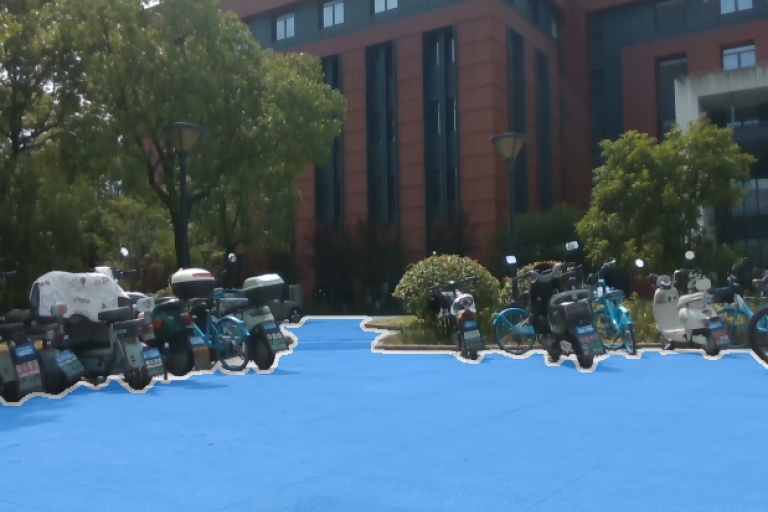}
\end{subfigure} \\
\begin{subfigure}[b]{0.30\linewidth}
    \includegraphics[width=\linewidth]{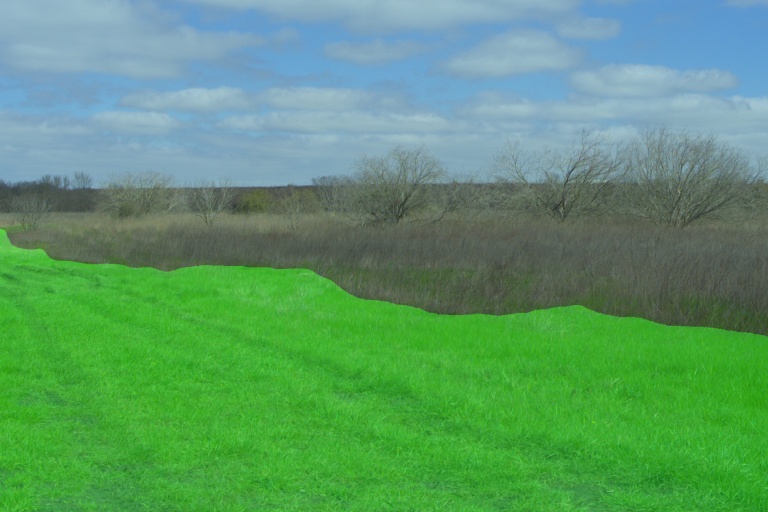}
\end{subfigure} &
\begin{subfigure}[b]{0.30\linewidth}
    \includegraphics[width=\linewidth]{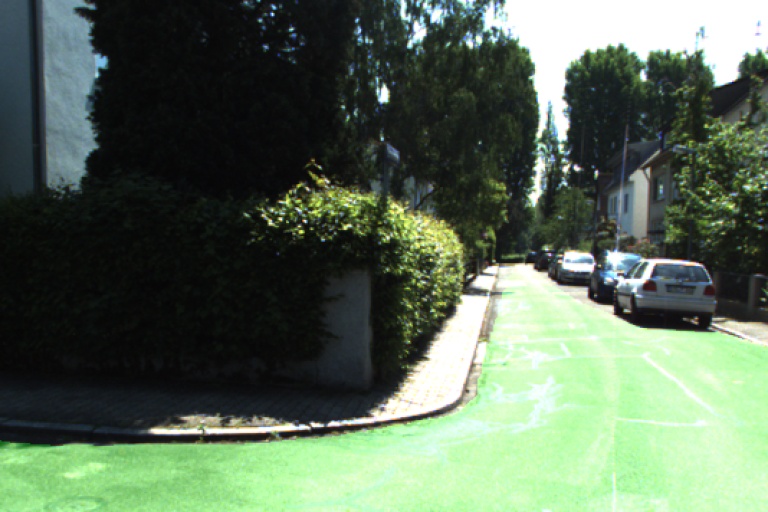}
\end{subfigure} &
\begin{subfigure}[b]{0.30\linewidth}
    \includegraphics[width=\linewidth]{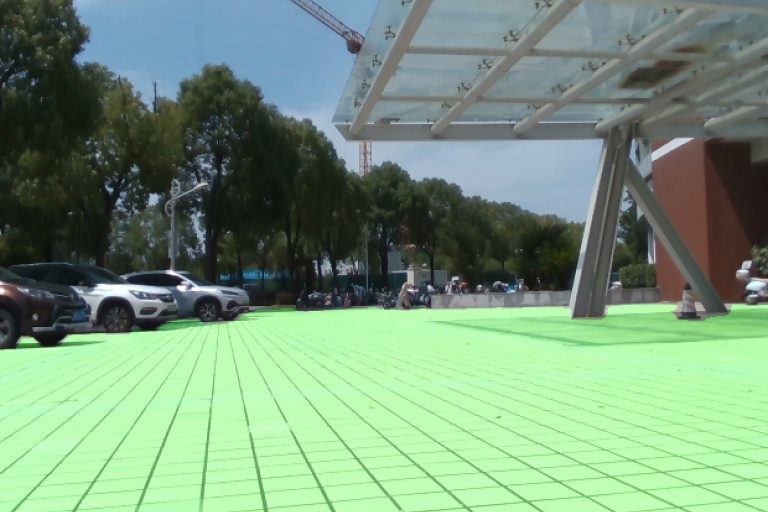}
\end{subfigure} \\
\end{tabular}
\caption{We present a multimodal self-supervised framework for traversability labeling and estimation in different scenes~\cite{Jiang_RELLIS3D,liao2022kitti}. Top: Collected data with projected footprints and LiDAR information overlaid on the image. Middle: Traversability labels automatically generated by the annotation pipeline. Bottom: Results of our multimodal traversability estimation network trained with self-supervised labels.}

\label{fig:main_figure}
\end{figure}

To address these limitations, we propose a multimodal self-supervised framework for traversability labeling and estimation that leverages both semantic and geometric information. The proposed framework consists of two key phases.
In the traversability labeling phase, we utilize geometric cues from LiDAR to identify hazardous regions without requiring physical traversal. In parallel, we leverage pretrained visual foundation models to obtain reliable positive regions. Positive and negative samples are selected via farthest point sampling and used as prompts for SAM to generate dense traversability labels. Meanwhile, we retain sparse LiDAR labels for refined supervision. In the traversability estimation phase, we employ a dual-stream network where semantic and geometric features are extracted independently in the encoder and fused in the decoder. To address potential errors in the automatic labeling pipeline, we apply a loss correction strategy guided by the sparse LiDAR labels, enhancing training robustness.

Our contributions are as follows:
\begin{itemize}
    \item We propose an automatic traversability labeling pipeline that leverages footprint, camera, and LiDAR data to generate pixel-wise traversability labels without human supervision. 
    \item We design a dual-stream network to capture diverse traversability patterns. In addition, we utilize sparse LiDAR labels to mitigate errors introduced by the automatic traversability labeling pipeline.
    \item We validate our method on diverse environments (off-road, urban, and campus), demonstrating improved annotation quality and state-of-the-art traversability estimation. Real-world deployment further highlights its effectiveness for downstream tasks such as planning and navigation.
\end{itemize}

\section{RELATED WORKS}

\subsection{Traversability Estimation}
Existing traversability estimation methods can be broadly categorized into three types: rule-based, segmentation-based, and self-supervised approaches.

Rule-based traversability estimation methods~\cite{Wermelinger_Navigation,Fan2021STEPST} typically rely on geometric features to characterize the terrain surfaces. Subsequent approaches~\cite{jian2022putn,xue2023traversability} often employ Bayesian inference or Gaussian processes to generate dense elevation maps that assist navigation~\cite{leininger2024gaussian,xie2023real}. However, these methods inherently suffer from regression errors and sensor noise, and struggle to handle flat but hazardous terrains, such as snowfields or puddles.

Semantically supervised segmentation methods~\cite{shaban2022semantic, guan2022ga} have incorporated human priors to heuristically assign cost values to different terrain regions. However, these methods are heavily dependent on manual terrain priors and often fail to generalize to unseen environments.

Self-supervised approaches~\cite{gherold2025self,cho2024learning,frey23fast} are promising directions, as they enable training directly from driving data without the need for human annotations.~\cite{seo2023learning,xue2023contrastive} utilize projected footprints as positive labels and adopt techniques from PU learning or semi-supervised learning to learn traversability embeddings.

We believe self-supervised learning holds significant potential for future development due to its low-cost nature and strong ability to generalize across diverse scenarios. Despite the clever designs for embedding space learning in existing studies, their limited access to non-traversable labels constrains the model capacity. We emphasize that learning the features of non-traversable areas is crucial, as these areas pose safety risks to the vehicle. This motivates us to explore additional priors to enrich the supervision and enhance the robustness of the model.

\subsection{Automatic Traversability Labeling}

With breakthroughs in large-scale pre-trained models in the visual domain~\cite{wang2025salt}, increasing efforts have been made to explore how these models can be leveraged for automatic traversability labeling. Kim et al.~\cite{kim2024learning} utilize future footprints with SAM to annotate traversable areas, but the short temporal range limits spatial coverage.
TADAP~\cite{alamikkotervo2024tadap} uses DINOv2 features and CRF refinement to propagate footprint semantics. However, its effectiveness relies on a high degree of visual homogeneity within traversable regions, limiting its adaptability to diverse and complex scenes.

These methods mainly emphasize positive footprint regions and often neglect the geometric cues. Some approaches~\cite{chen2024learning, gherold2025self} incorporate geometric heuristics but suffer from inconsistent labels. Alamikkotervo et al.~\cite{alamikkotervo2024trajectory} leverage interpolation to extend geometric labels but their method assume that the driving region is lower in the middle and higher on the edges, which limits its generalization ability. In contrast, our framework introduces minimal prior assumptions, generalizes well across diverse environments, and generates dense and consistent traversability labels.

\subsection{Multimodal Traversability Estimation Network}


Semantic and geometric information are both essential for robust traversability estimation. FtFoot~\cite{jeon2024follow} leverages RGB-D data to predict normals, which are then used to guide image feature extraction, achieving the fusion of different modalities. However, we believe that semantic and geometric information in traversability estimation tasks is not highly coupled. For instance, a lawn, even though relatively flat geometrically, is a non-traversable area owing to its semantic nature. This prompts us to design a decoupled method for extracting multimodal information. Similar to our approach, GrASPE~\cite{weerakoon2023graspe} extracts image and point cloud features separately, and employs graph neural networks to fuse these with trajectory and velocity data for future trajectory safety assessment. Differing from their work, our method performs traversability estimation at the granularity of each pixel.

\section{METHODS}
\begin{figure*}[htbp!]
\centering
\includegraphics[width=0.8\textwidth]{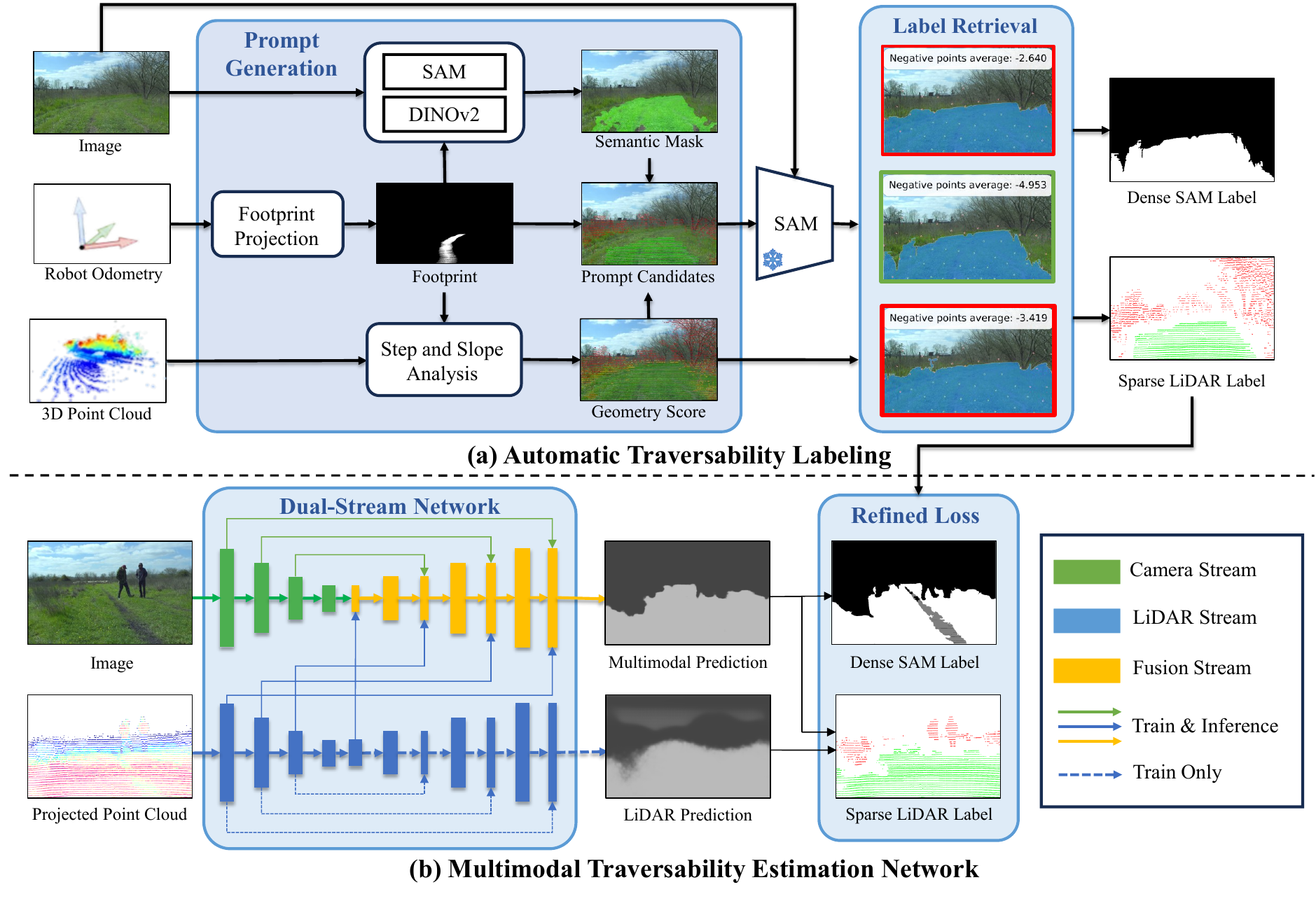}
\caption{Overall framework. (a) The top illustrates our automatic labeling pipeline. Semantic and geometric cues are used to generate prompt candidates for SAM.
Geometric negative points are then used to guide the final label retrieval. (b) The bottom shows the dual-stream network training with the label generated by our automatic labeling pipeline. The masked regions in dense SAM labels are shown in gray, and sparse LiDAR labels provide auxiliary supervision.}
\label{fig:pipeline}
\vspace{-5mm}
\end{figure*}

Our objective is to self-supervisedly train a multimodal traversability estimation network capable of perceiving both semantic and geometric risks in diverse terrains and scenes without requiring human annotations. Specifically, given an image \( I \) and LiDAR data \( L \), the network learns to predict the pixel-wise traversability probabilities. The overall framework of our approach is shown in \Fref{fig:pipeline}. 

\subsection{Automatic Traversability Labeling}
Our automatic annotation pipeline comprises two main stages. The first stage is the \textbf{prompt generation} stage, which extracts footprint information, expands it using semantic and geometric priors, and finally fuses both to construct the refined prompt candidates. The second stage is the \textbf{label retrieval} stage, which selects the most suitable result from multiple SAM outputs.

\subsubsection{Initial Data Acquisition and Footprint Projection}
Our data acquisition method is applicable to various mobile robots equipped with LiDAR and cameras. The process begins by driving the robot through areas perceived as traversable under human cognition. 

Thanks to mature SLAM systems~\cite{xu2021fast,deng2025mne}, robust pose estimation enables accurate footprint projection. Following~\cite{jeon2024follow}, we predefine the sensor height $h$ and robot footprint radius $r$. The position of each $i$-th LiDAR frame in the global coordinate system is projected onto the ground plane to obtain ${^{G}p_{L,i}^{\prime}(x, y, z - h)}$. The corresponding footprint region can be represented by the set ${^{G}P_{i}^{f}} = U({^{G}p_{L,i}^{\prime}(x, y, z - h)}, r)$, where $U$ denotes the neighborhood centered at $p$ on the $xoy$ plane with a radius $r$.
Thus, the footprint regions over $N$ frames are given by:
${^{G}P_{0:N}^{f}} = { {^{G}P_{0}^{f}} \cup \cdots \cup {^{G}P_{N}^{f}} }$.
We project the footprints from $n$ future frames onto the current image frame:
\begin{equation}\label{equation:footprint_projection}
p^f_{i} = K  {^{C}T_{L}} {^{L}T_{G,i}}  {^{G}P^f_{i:i+n}},
\end{equation}
where $K$ and $^{C}T_{L}$ represent intrinsic camera calibration matrix and extrinsic transformation matrix respectively; $^{L}T_{G,i}$ represent the LiDAR pose in the global coordinate system at the current frame; and $p^f_{i}$ is the set of projected footprint pixels.

We propose that incorporating both a longer horizon and surrounding radius ensures comprehensive coverage of the footprint, capturing rich information of traversable regions. The projected footprint regions are also mapped into the current BEV space to support the geometry score computation.



\subsubsection{Incorporating Semantic and Geometric Priors}
\begin{figure}[htbp!]
\centering
\vspace{0.1cm}
\begin{tabular}{*{2}{c@{\hspace{6px}}}}
\begin{subfigure}[b]{0.20\textwidth}
    \includegraphics[width=\textwidth]{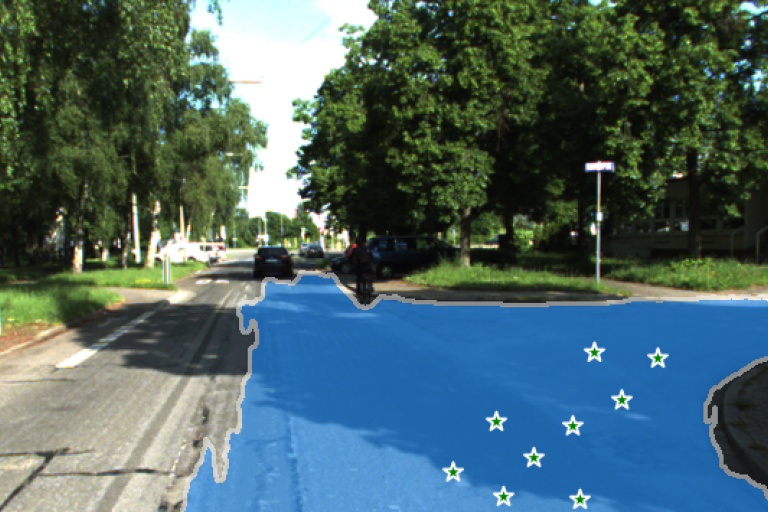}
    \caption{Over-segment}
    \label{fig:sam_mistake_a}
\end{subfigure} &
\begin{subfigure}[b]{0.20\textwidth}
    \includegraphics[width=\textwidth]{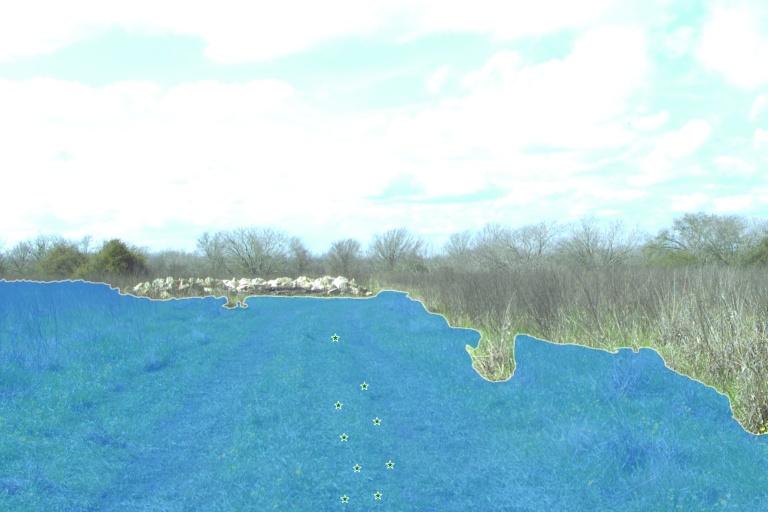}
    \caption{Under-segment}
    \label{fig:sam_mistake_b}
\end{subfigure} \\
\begin{subfigure}[b]{0.20\textwidth}
    \includegraphics[width=\textwidth]{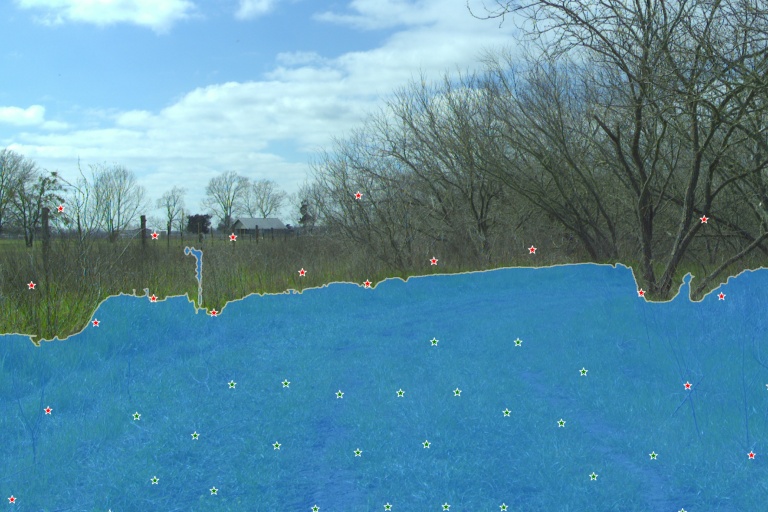}
    \caption{Max-score mask}
    \label{fig:sam_mistake_c}
\end{subfigure} &
\begin{subfigure}[b]{0.20\textwidth}
    \includegraphics[width=\textwidth]{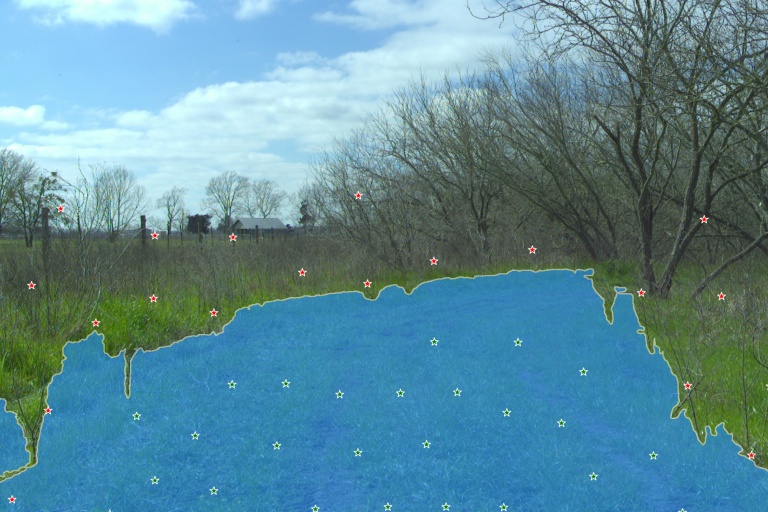}
    \caption{Our chosen mask}
    \label{fig:sam_mistake_d}
\end{subfigure}
\end{tabular}
\caption{Limitations of the existing SAM-based method~\cite{kim2024learning}. (a) and (b) show that using only positive examples can yield suboptimal masks due to structural boundaries or texture similarity. (c) presents the max-score mask, while (d) displays the improved result obtained through our retrieval strategy.}
\label{fig:sam_mistake}
\vspace{-2mm}
\end{figure}

Using only the footprint projection as the point prompts introduces two significant problems, as illustrated in Fig.~\ref{fig:sam_mistake_a} and Fig.~\ref{fig:sam_mistake_b}. First, SAM often results in over-segmentation, particularly around the boundary structures of traversable areas. Second, SAM fails to effectively distinguish ambiguous obstacles such as bushes. To address these issues, we incorporate semantic and geometric priors to provide more comprehensive information for point prompts.

For the semantic branch, we combine SAM-based and similarity-based region expansion to mitigate their respective limitations. SAM offers strong discrimination but suffers from boundary artifacts, whereas similarity-based expansion is more generalizable but prone to confusion in texture-similar areas. Specifically, we apply Farthest Point Sampling~\cite{qi2017pointnet} on the footprint projection to generate prompts for SAM to obtain SAM-based mask and adopt the method in~\cite{alamikkotervo2024tadap}, which utilizes DINOv2 pretrained features to obtain the similarity-based mask. The final semantic mask is computed as the intersection of the SAM and DINOv2 masks. If the SAM mask is significantly smaller, indicating potential over-segmentation, we fallback to the DINOv2 result.

\begin{equation}
\mathcal{M}_{\mathrm{sem}} =
\begin{cases}
\mathcal{M}_{\mathrm{DINOv2}}, & \text{if } |\mathcal{M}_{\mathrm{SAM}}| \leq \lambda_a \cdot |\mathcal{M}_{\mathrm{DINOv2}}|, \\
\mathcal{M}_{\mathrm{SAM}} \cap \mathcal{M}_{\mathrm{DINOv2}}, & \text{otherwise,}
\end{cases}
\end{equation}
where $|\mathcal{M}_{\mathrm{SAM}}|$ and $|\mathcal{M}_{\mathrm{DINOv2}}|$ denote the areas of the SAM and DINOv2 masks, respectively, and $\lambda_a$ is the area scaling factor.

For the geometric branch, we adopt two fundamental geometric descriptors: step and slope. We compute step and slope~\cite{leininger2024gaussian} directly on each point in a single-frame LiDAR scan, rather than on grid-level representations. This avoids the need for multi-frame accumulation and mitigates ghosting from dynamic objects. Analyzing geometric attributes at the point level also reduces grid discretization errors.

After computing geometric attributes for each point, we fit the score distribution within the BEV footprint and define the $\tau$-th percentile of the step and slope scores as the geometric baseline $b_g$. This baseline captures the vehicle’s traversability across varying terrain conditions, enabling better adaptation to environmental changes. Geometric features are subtracted from this baseline and normalized to [0, 1] via exponential mapping. The normalized step and slope scores are then fused through weighted averaging to yield a geometric traversability score for each LiDAR point.
\begin{equation}
T_{\mathrm{geo}}(p) = \sum_{g \in \{\mathrm{step}, \mathrm{slope}\}} 
    w_g \cdot \exp \left( -\frac{\max \left( 0, g(p) - b_g \right)}{c_g} \right),
\end{equation}
where $w_g$ denotes the attribute weight, $g(p)$ represents the geometric attribute for each point, and $c_g$ is the attribute scaling factor.


\subsubsection{Refined Prompt Candidates Construction}
After gaining a richer understanding of safe and hazardous regions in the image, we utilize it to construct the refined prompt candidates.

For positive prompts, we first select LiDAR points with the geometric traversability score exceeding the high traversability threshold and project them onto the image. We then verify that these points also align with the results obtained from the semantic branch.
\begin{equation}
\begin{aligned}
\mathcal{P}^{+}_{\mathrm{cand}} = \big\{ p = \Pi_{\mathrm{cam}}(q) \;\big| \;
& p \in {\mathcal{M}}_{\mathrm{sem}}, \\
& T_{\mathrm{geo}}(q) > \tau_{\mathrm{pos}}, \forall q \in \mathcal{P}_{\mathrm{LiDAR}} \big\}
\end{aligned}
\end{equation}
where  $\Pi_{\mathrm{cam}}$ denotes the projection function, $\tau_{\mathrm{pos}}$ is the poisitive traversability threshold, and ${\mathcal{M}}_{\mathrm{sem}}$ and $T_{\mathrm{geo}}(q)$ are drawn from semantic and geometric branch above.

To account for occlusions in the image space, we exclude points near LiDAR points with the traversability score below the low traversability threshold.
\begin{equation}
\begin{aligned}
\mathcal{P}^{+} = \big\{ p \in \mathcal{P}^{+}_{\mathrm{cand}} \;\big|\;
&  \; T_{\mathrm{geo}}(q) < \tau_{\mathrm{neg}}, \\
& \| p - \Pi_{\mathrm{cam}}(q) \|_2 > \delta,\forall q \in \mathcal{P}_{\mathrm{LiDAR}} \big\}.
\end{aligned}
\end{equation}

For negative prompts, recognizing that SAM effectively distinguishes obvious obstacles, such as tall buildings or trees, we expand the SAM mask generated by positive prompts through a dilation operation. $\mathrm{Dilate}(\cdot)$ indicates increasing the mask's boundary to include nearby regions. Then, we select non-traversable LiDAR points within the dilated mask as negative prompt candidates. This design encourages negative prompts to cover ambiguous regions where SAM might be confused. We also filter out negative points within the footprint to avoid potential errors and improve robustness.
\begin{equation}
\widetilde{\mathcal{M}}_{\mathrm{SAM}} = \mathrm{Dilate}\left( \mathcal{M}_{\mathrm{SAM}} \right),
\end{equation}
\begin{equation}
\mathcal{P}^{-} = \left\{ p = \Pi_{\mathrm{cam}}(q) \ \middle|\ T_{\mathrm{geo}}(q) < \tau_{\mathrm{neg}},\ p \in \widetilde{\mathcal{M}}_{\mathrm{SAM}} \right\},
\end{equation}
where ${\mathcal{M}}_{\mathrm{SAM}}$ is the mask prompted only by the footprint, and $\tau_{\mathrm{neg}}$ is the low traversability threshold.

\subsubsection{Label Retrieval}
Finally, we apply Farthest Point Sampling \cite{qi2017pointnet} to select a fixed number of positive and negative prompts and input them to SAM. However, as illustrated in Fig.~\ref{fig:sam_mistake_c}, simply choosing the mask with the highest score from three candidate masks may produce the wrong label. Therefore, our retrieval strategy is to select the candidate that yields the lowest average activation score over the negative points to suppress responses in non-traversable regions.
\begin{equation}
M^* = \arg\min_{M \in \mathcal{M}_{\mathrm{final}}} \frac{1}{|\mathcal{P}^{-}|} \sum_{p \in \mathcal{P}^{-}} S_{\mathcal{}{M}}(p),
\end{equation}
where $\mathcal{M}_{\mathrm{final}}$ are multiple SAM masks generated by our refined prompts, and $S_{\mathcal{}{M}}(p)$ is the pixel-wise activation score in each mask. 

The selected mask is used as our dense SAM label. Notably, we retain the positive and negative prompt candidates as the sparse LiDAR label. Since masks generated by SAM may not respond to every prompt, the sparse LiDAR label provides additional supervision signals to improve training robustness. These labels are then leveraged to train our multimomdal traversability estimation network.

\subsection{Multimodal Traversability Estimation Network}

\subsubsection{Network Architecture Design}
Our design is grounded in the nature of traversability: if either the semantic or geometric branch identifies an area as non-traversable, the final decision should be non-traversable. Therefore, instead of frequent feature-level interactions between the two modalities at various stages, we aim to independently encode the semantic and geometric information and fuse them at the decoder stage.

Based on this intuition, we design a dual-stream network where each branch extracts semantic and geometric features respectively and produces multi-scale hierarchical representations. In the decoder, features from both modalities are concatenated and progressively upsampled to produce a traversability probability map at the original image resolution.



\subsubsection{Network Inputs and Loss Function}
Following the approach in~\cite{tan2024epmf}, in the LiDAR branch, we project the LiDAR point cloud onto the 2D plane to form LiDAR stream input with dimensions $R \in \mathbb{R}^{5 \times H \times W}$, containing five channels: $(x, y, z, r, d)$ representing coordinates, reflectance intensity, and distance, respectively. The camera branch receives the RGB image with dimensions $R \in \mathbb{R}^{3 \times H \times W}$. The camera encoder is pretrained on ImageNet. To effectively leverage LiDAR information, we also pretrain the LiDAR encoder using sparse LiDAR labels.

Due to inherent limitations of SAM and conservative retrieval strategy, the dense SAM labels may contain errors. To mitigate this, we dilate the sparse LiDAR labels and mask out regions in the SAM output that conflict with any dilated LiDAR annotations, leveraging the more reliable geometric priors provided by LiDAR information.

The refined loss comprises two components: Lovász loss \cite{berman2018lovasz}, applied to dense SAM labels with regions masked out; and Cross-Entropy (CE) loss, applied directly to the sparse LiDAR labels to strengthen supervision in areas where SAM predictions are inaccurate. The combined loss is formulated as follows:

\begin{equation}
\mathcal{L} = \mathrm{Lov\acute{a}sz}(\hat{y}, y_{\mathrm{SAM\text{-}masked}}) + \lambda_s \cdot\mathrm{CE}(\hat{y}, y_{\mathrm{LiDAR\text{-}sparse}}),
\end{equation}
where $\hat{y}$ denotes the network's multimodal prediction, $y_{\mathrm{SAM\text{-}masked}}$ represents the masked dense SAM labels, $y_{\mathrm{LiDAR\text{-}sparse}}$ refers to the sparse LiDAR labels, and $\lambda_s$ is the auxiliary loss weight.

\section{Experiments}

\subsection{Datasets}\label{datasets}
To comprehensively assess our approach, we consider three representative types of environments: 

1) RELLIS-3D~\cite{Jiang_RELLIS3D}: We use the public RELLIS-3D dataset, a multimodal off-road dataset that includes data from a Basler camera and an Ouster 64-beam LiDAR. 
This dataset contains 13556 frames. We use sequences 0, 2, 3, and 4 for training and validation, and sequence 1 for testing.

2) KITTI-360~\cite{liao2022kitti}: We also employ the widely used KITTI-360 dataset for structured urban driving scenarios. This multimodal dataset covers a range of urban road scenes.
This dataset contains 76527 frames. Sequences 0, 2, 3, 4, 5, 6, and 7 are used for training and validation, while sequences 9 and 10 are reserved for testing.

3) Campus Dataset: We construct a custom dataset collected by our wheelchair robot platform, which is equipped with a Realsense D435i RGB-D camera and a RS-LiDAR-16. The Campus dataset includes roads, sidewalks, and park paths. The dataset contains 8407 frames. One manually labeled path is used for testing, and the remaining sequences are used for training and validation.

\subsection{Experimental Setup}\label{setup}
We conduct two experiments to evaluate the effectiveness of our proposed framework. For automatic traversability labeling evaluation, we compare our method against the following baselines: (1) EVAA~\cite{kim2024learning}, a SAM-based approach that utilizes future footstep locations over a three-second horizon as prompts; (2) RALCF~\cite{alamikkotervo2024trajectory}, a method that separately annotates traversable areas using both camera and LiDAR data, and combines them through mean fusion and CRF. For traversability estimation evaluation, we compare our method with state-of-the-art self-supervised methods: (1) EVAA, which finetunes a lightweight segmentation model using the above annotations, applying L2 loss to positive labels and a small-weight L1 loss to negative labels. (2) FtFoot~\cite{jeon2024follow}, which relies on sparse footstep supervision and uses projection depth to guide image convolution; (3) V-STRONG~\cite{jung2024v}, which integrates SAM with ViT and contrastive learning.


We employ standard evaluation metrics including Intersection over Union (IoU), False Positive Rate (FPR), False Negative Rate (FNR), Average Precision (AP), F1-score (F1), Precision (Pre), and Recall (Rec). We report metrics at the best threshold that achieves the highest F1-score.

\subsection{Implementation Details}\label{implement}
For the automatic traversability labeling, in RELLIS-3D, due to the uneven terrain, we set $c_{\mathrm{step}}$ as 0.1 and $c_{\mathrm{slope}}$ as 0.15, with corresponding weights $w_{\mathrm{step}}$ and $w_{\mathrm{slope}}$ both assigned 0.5. Given the high image resolution and complex scene, we sample 25 positive and negative prompts to cover more area. In KITTI-360 and Campus dataset, we only consider the step attribute to detect low obstacles, setting $c_{\mathrm{step}}$ as 0.05, and sample 10 positive and negative prompts. The area scaling factor $\lambda_a$ is fixed at 0.8. $\tau$ is set to 80 to filter out outliers. $\tau_{\mathrm{pos}}$ and $\tau_{\mathrm{neg}}$ are assigned values of 0.9 and 0.1, respectively. The dilated kernel radius is set to 100 to encompass ambiguous regions while avoiding ineffective prompts.

The multimodal traversability estimation network employs a ResNet-34 backbone for the image encoder, while the LiDAR encoder adopts a SalsaNext architecture modified with sparse convolutions. We train for 10 epochs with an initial learning rate of 0.001. After a warm-up in the first epoch, the learning rate decays via a cosine schedule. The weight of the auxiliary loss is empirically set to $\lambda_s = 0.4$ following~\cite{feng2025sne}.

\subsection{Evaluation of Automatic Traversability Labeling}
\begin{figure}[h!]
	\centering
	\begin{tabular}{*{3}{c@{\hspace{1.8px}}}}
         \includegraphics[width=0.30\linewidth]{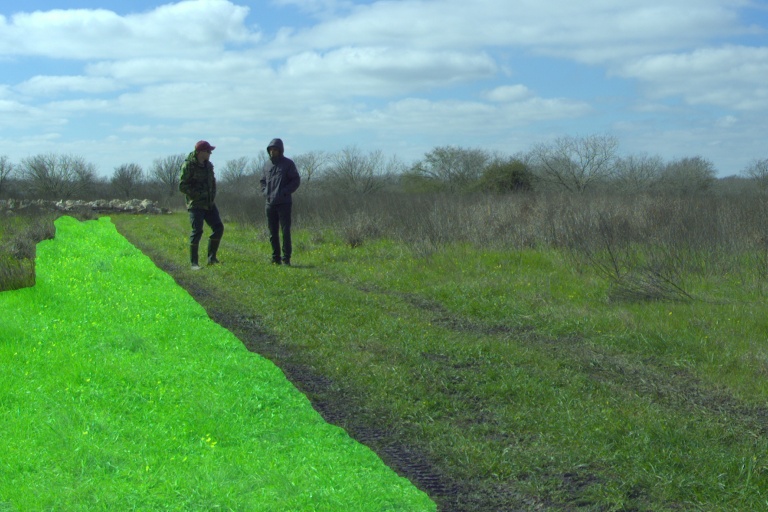} 
         & \includegraphics[width=0.30\linewidth]{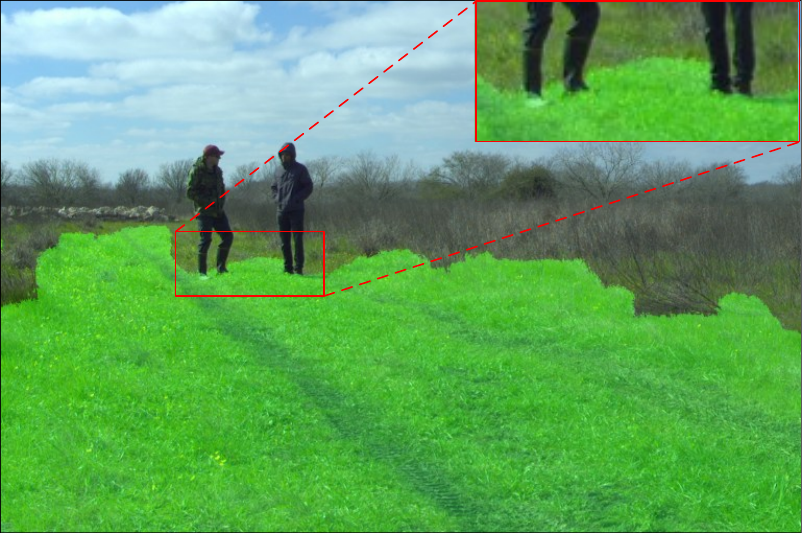} 
         & \includegraphics[width=0.30\linewidth]{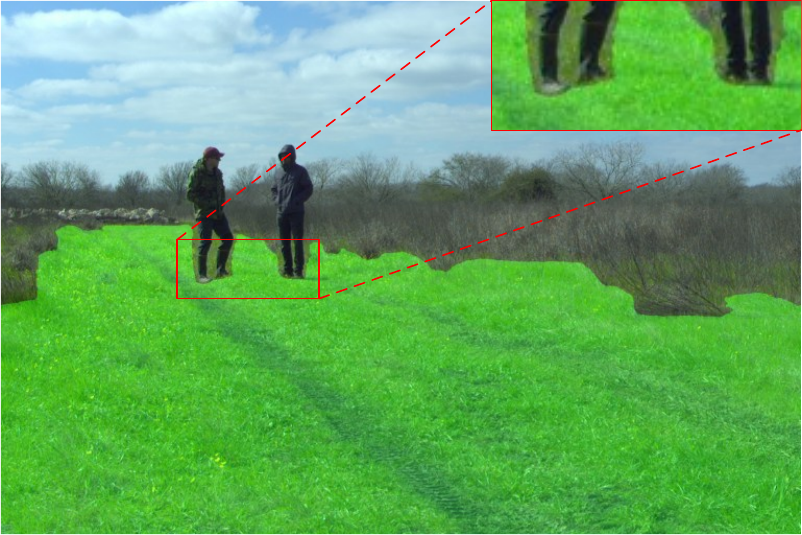} 
         \\
          \includegraphics[width=0.30\linewidth]{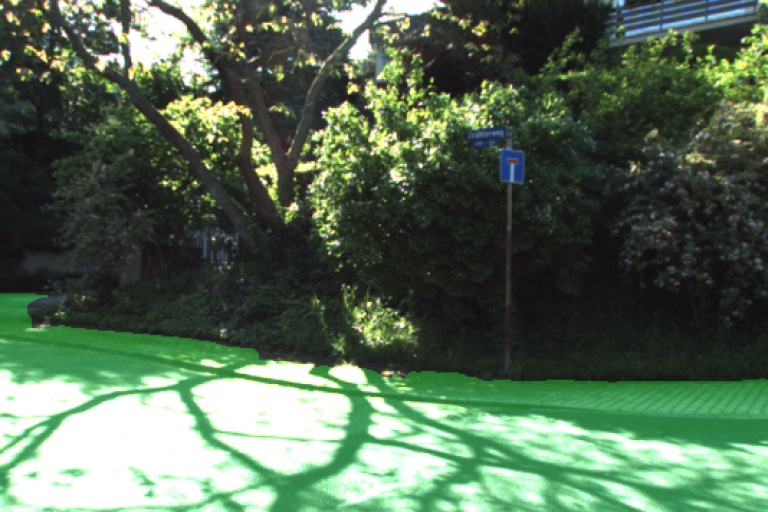} 
         & \includegraphics[width=0.30\linewidth]{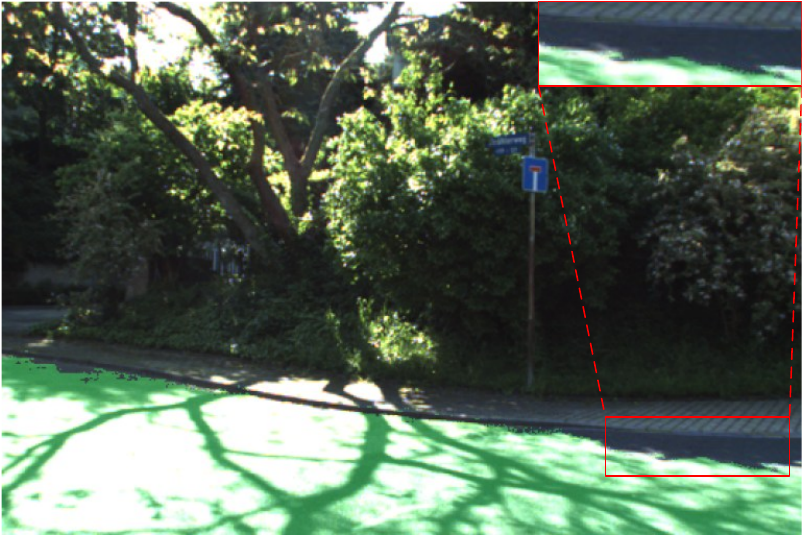} 
         & \includegraphics[width=0.30\linewidth]{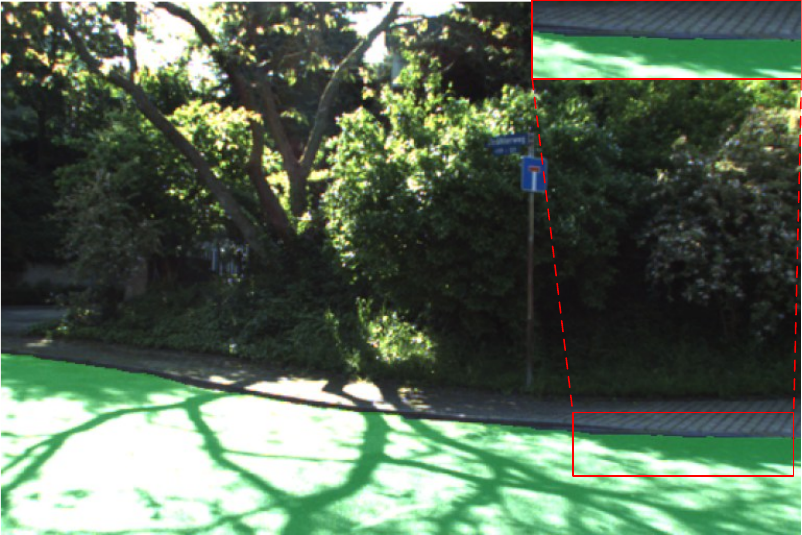} 
         \\
         
         \small{(a) EVAA} & \small{(b) RALCF} & \small{(c) Ours} 
	\end{tabular}
	\caption{Qualitative results of different methods for automatic traversability labeling. The first row is the results on Rellis-3D and the second row is the results on KITTI-360. The pixels colored in green are automatic annotation labels.}
	\label{fig:results_annotation}
\vspace{-2mm}
\end{figure}

The \Tref{tab:results_annotation} and \Fref{fig:results_annotation} present both the quantitative and qualitative results of our automatic traversability labeling method across various datasets.

Notably, on the RELLIS-3D dataset, EVAA exhibits a 4\% decrease on IoU  compared to KITTI-360. This is primarily due to the visual similarity between classes such as grass and bushes. And RALCF shows poor recall performance, as the similarity-based methods fail to sufficiently cover the entire traversable area.

As shown in the results, our approach consistently outperforms baselines by 4.1\% to 9.4\% on IoU and demonstrates stable performance across diverse environments. This improvement stems from our refined prompt and retrieval strategy, which enhances SAM's semantic understanding and significantly boosts both precision and recall. Although our method shows slightly lower recall than EVAA due to our conservative retrieval strategy aimed at reducing false positives, this limitation can be mitigated by incorporating refined supervision from sparse LiDAR labels.

\begin{table}[htbp!]
\centering
\caption{Quantitative results of different methods for automatic traversability labeling. Best results are in \textbf{bold}.}
\label{tab:results_annotation}
\resizebox{0.9\linewidth}{!}{%
\begin{tabular}{llcccc}
\toprule
Dataset & Method        & IoU & F1  & Pre   & Rec \\
\midrule
\multirow{3}{*}{RELLIS-3D} 
        & EVAA               & 0.803   & 0.891 & 0.866          & 0.917       \\
        & RALCF                  & 0.804      & 0.891     & 0.926          & 0.859         \\
        & Ours         & \textbf{0.879}  & \textbf{0.936} & \textbf{0.927} & \textbf{0.944}\\
\midrule
\multirow{3}{*}{KITTI-360} 
        & EVAA                      & 0.848 & 0.918    & 0.908          & \textbf{0.928}         \\
        & RALCF                         & 0.785    & 0.879   & \textbf{0.979} & 0.798           \\
        & Ours               & \textbf{0.883}   & \textbf{0.938}& {0.964}        & 0.914  \\
\bottomrule

\end{tabular}
}
\end{table}

\subsection{Evaluation of Traversability Estimation}
\begin{figure*}[htbp!]
	\centering
	\begin{tabular}{*{5}{c@{\hspace{1.8px}}}}
         \includegraphics[width=0.18\linewidth]{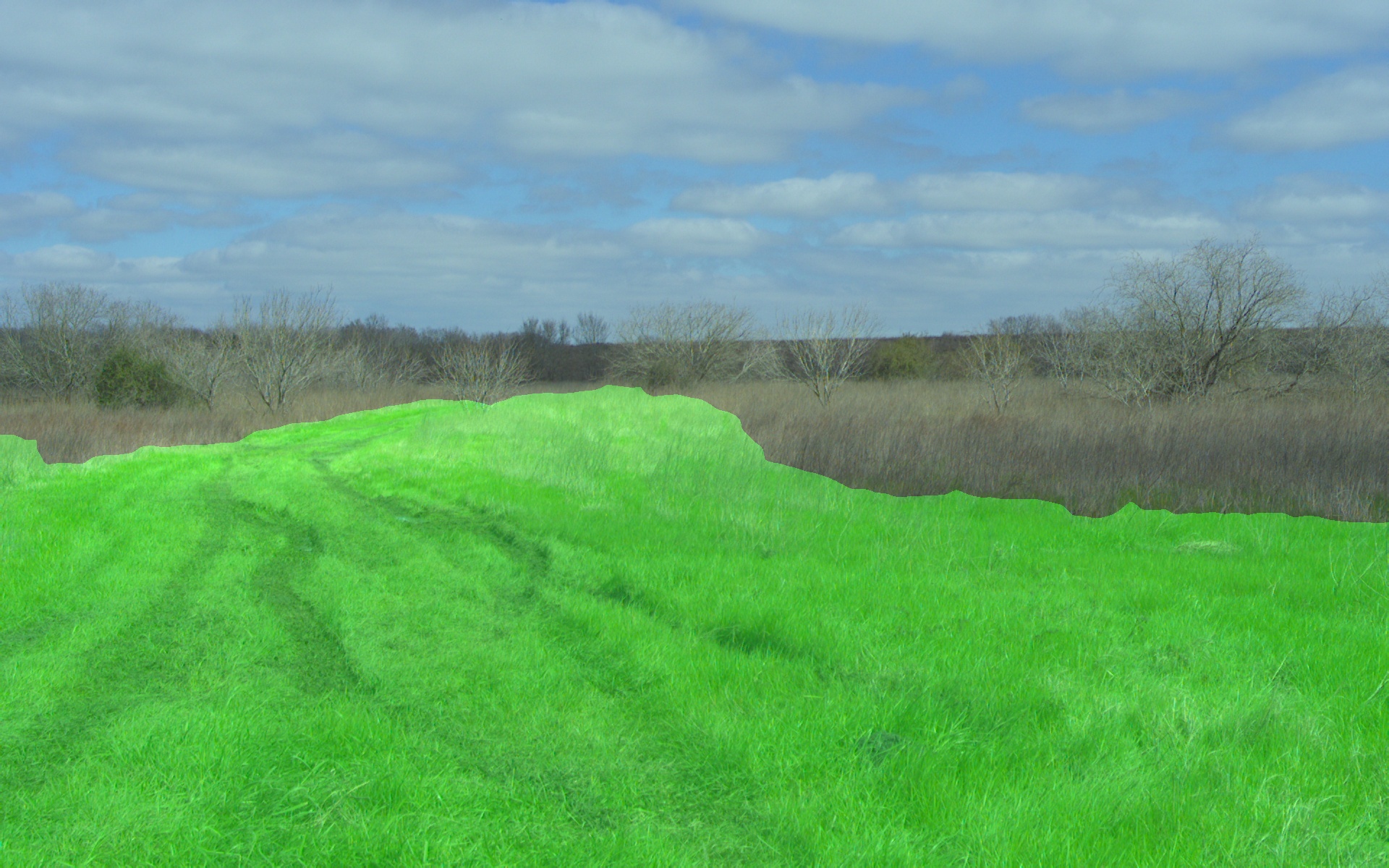} 
         & \includegraphics[width=0.18\linewidth]{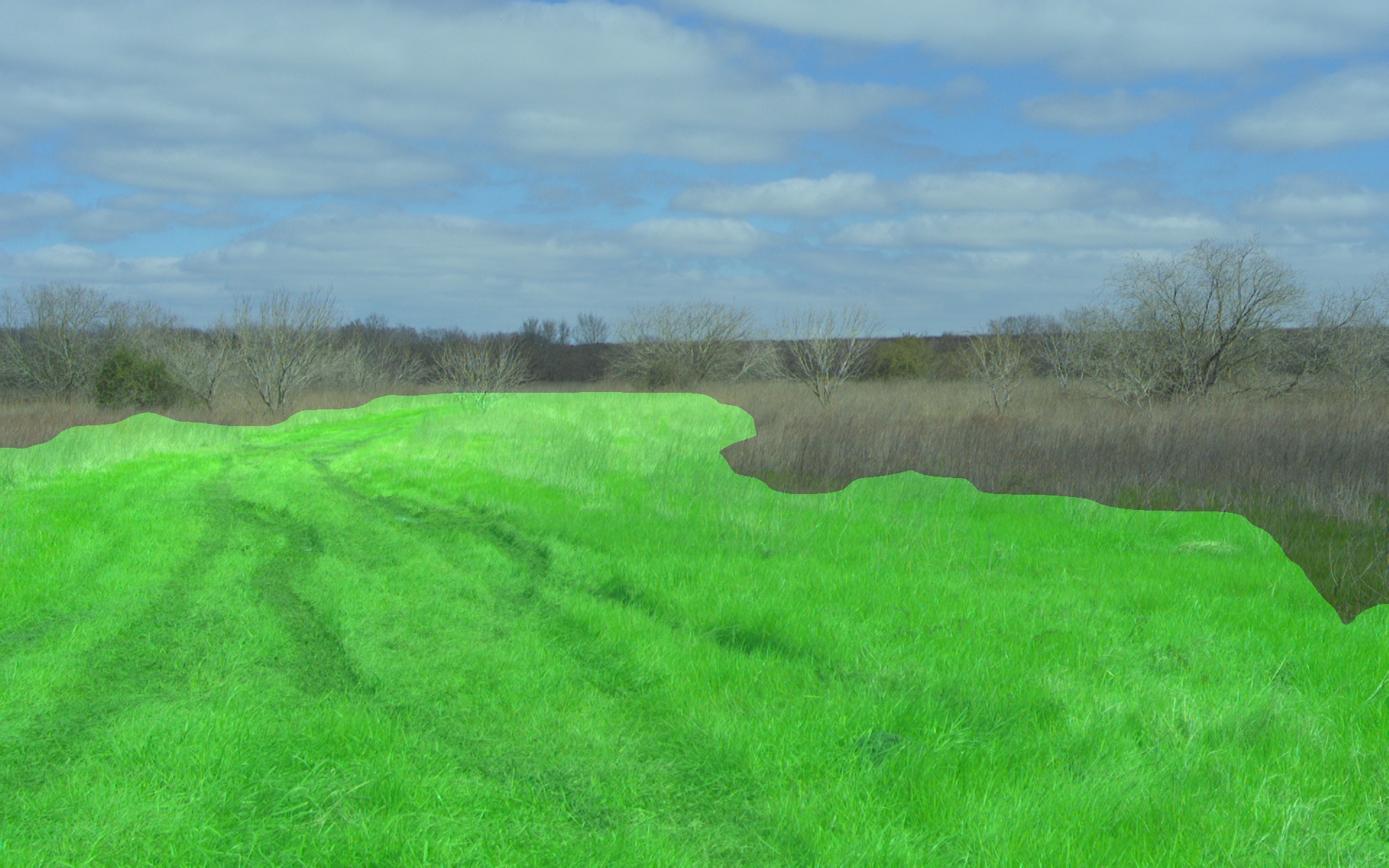} 
         & \includegraphics[width=0.18\linewidth]{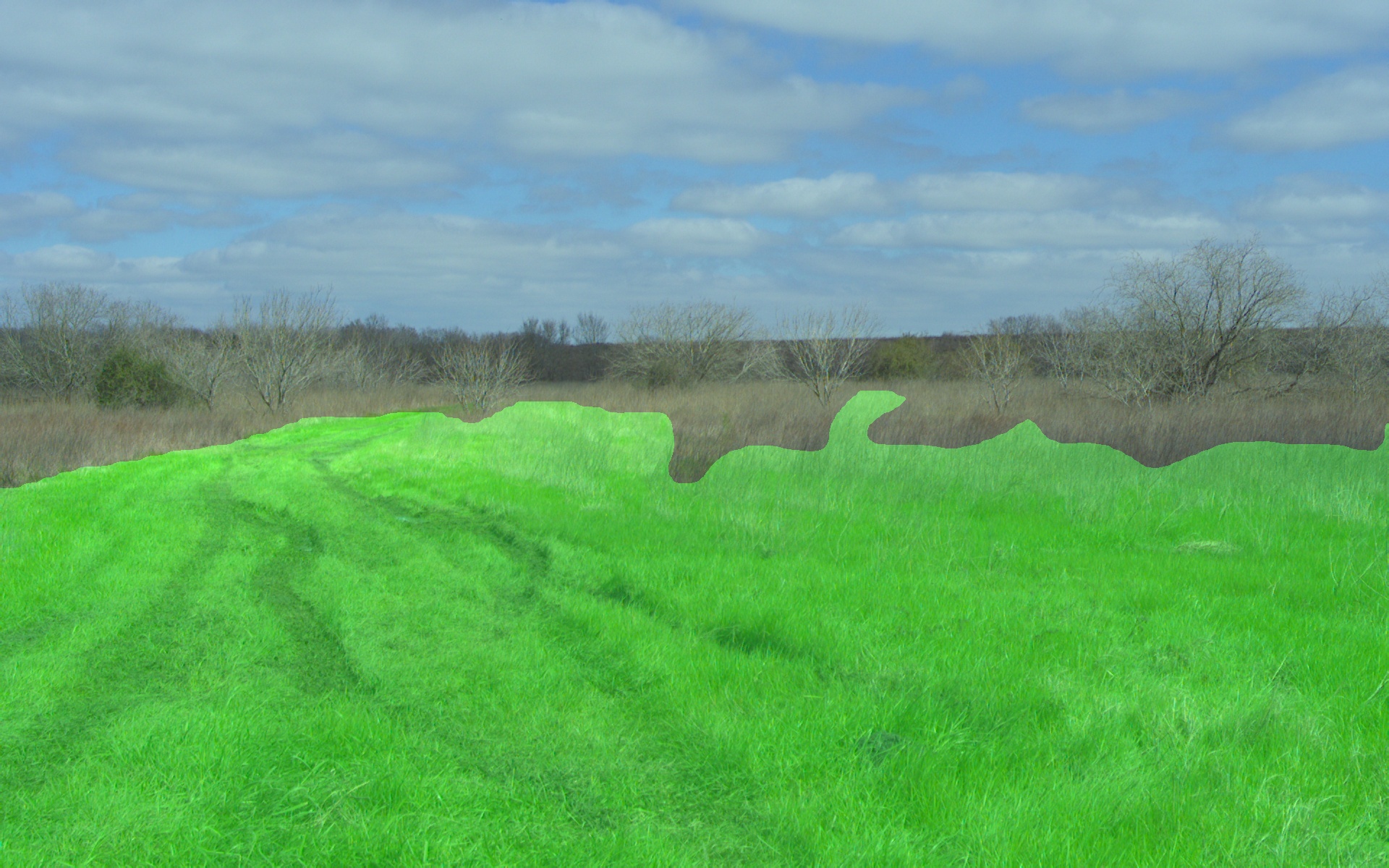} 
          & \includegraphics[width=0.18\linewidth]{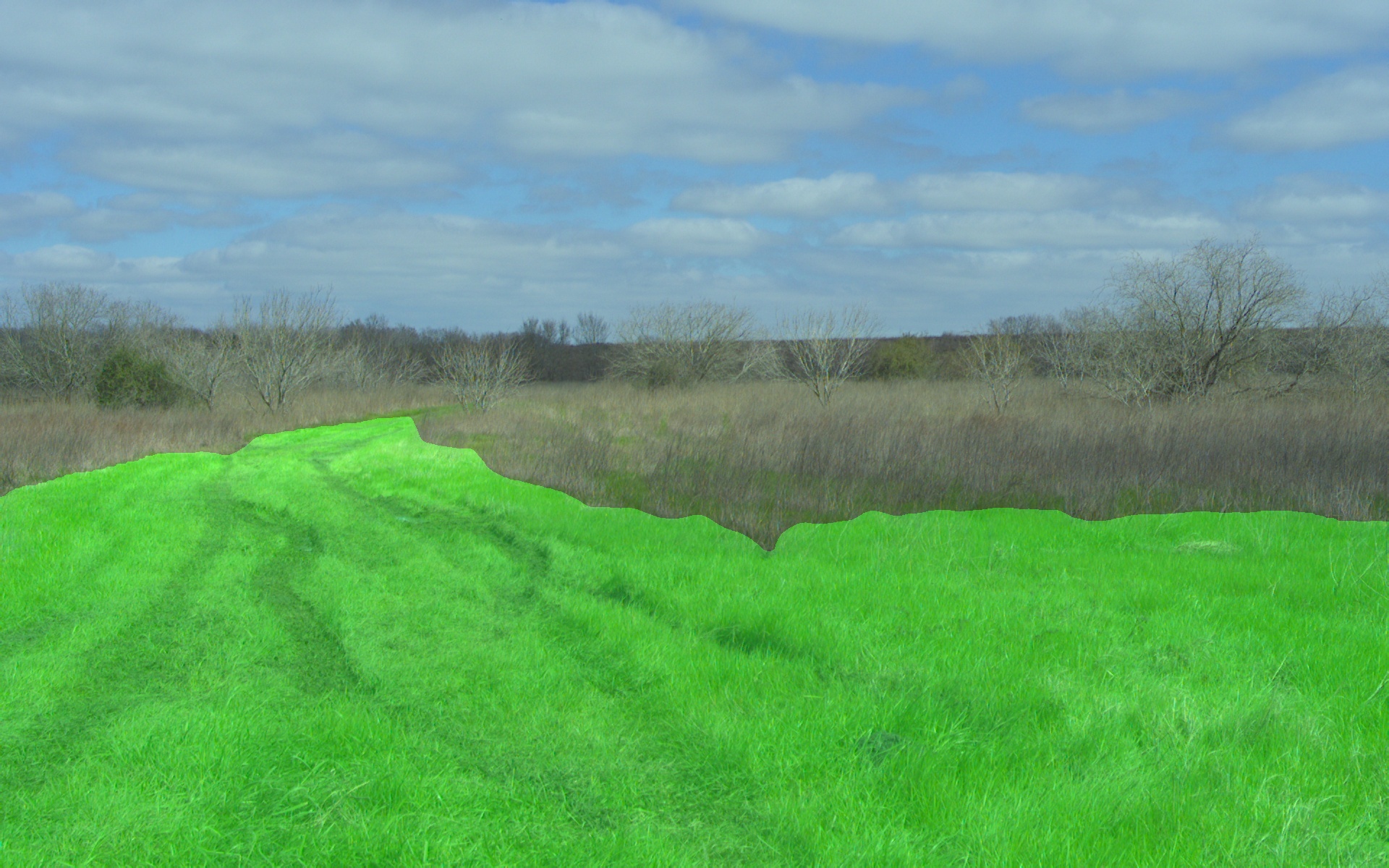} 
          & \includegraphics[width=0.18\linewidth]{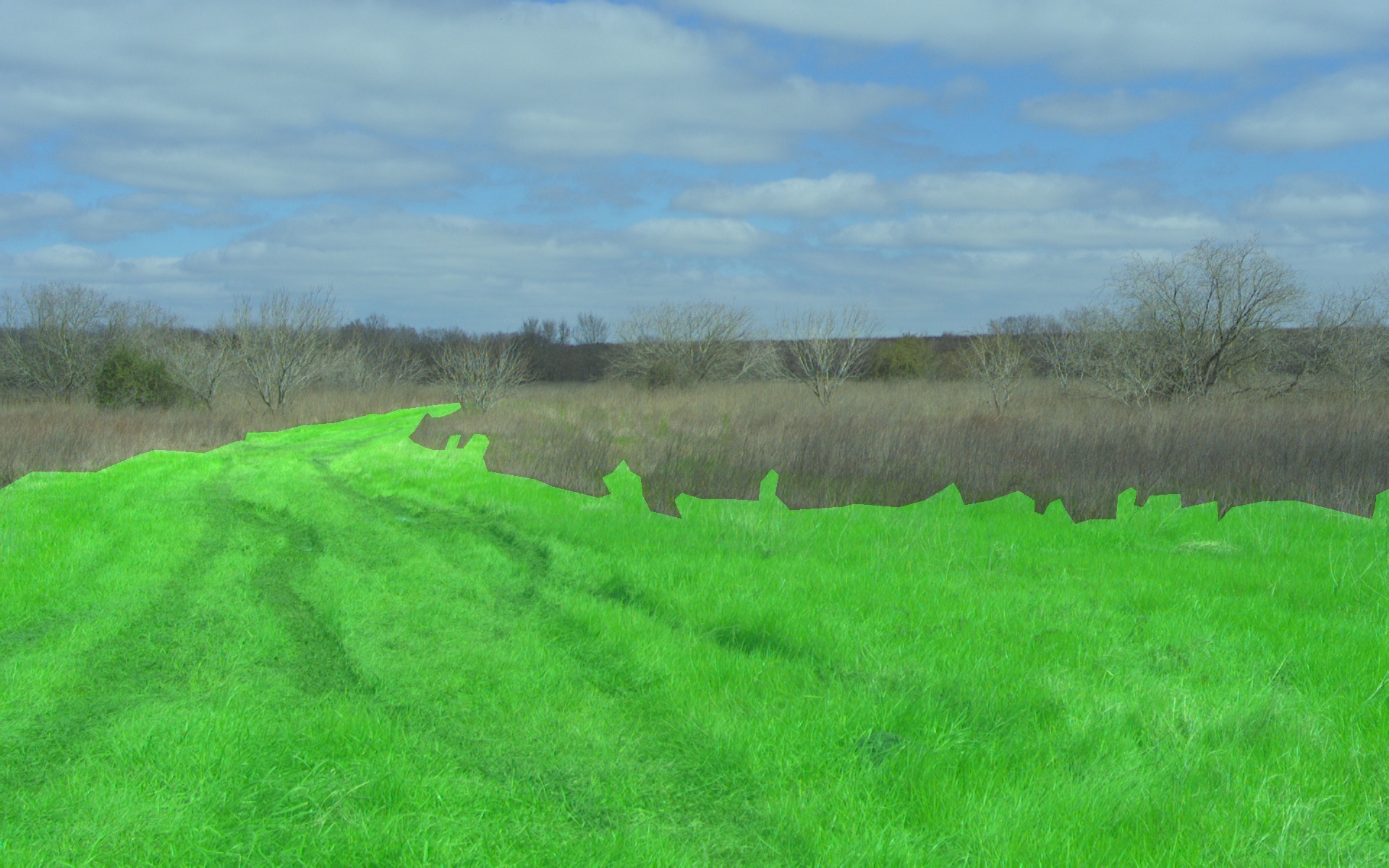} 
         \\
          \includegraphics[width=0.18\linewidth]{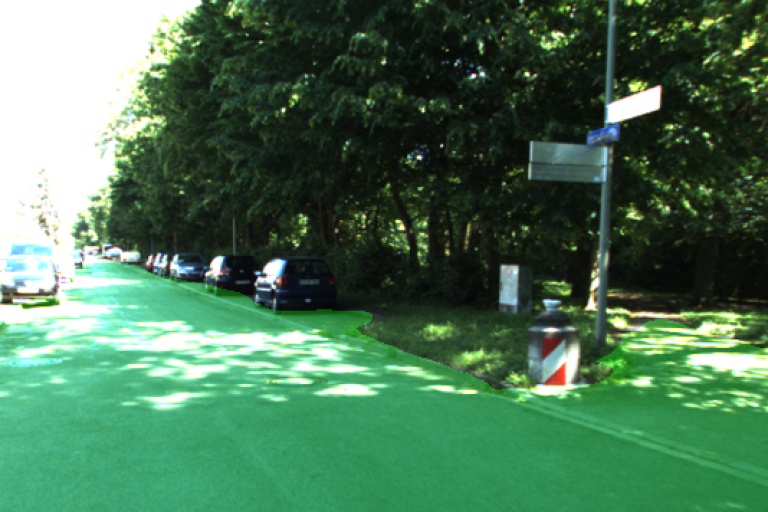} 
         & \includegraphics[width=0.18\linewidth]{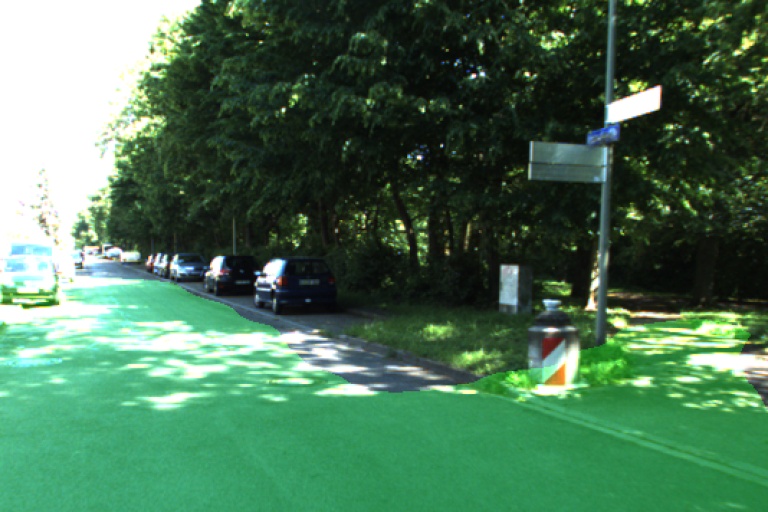} 
         & \includegraphics[width=0.18\linewidth]{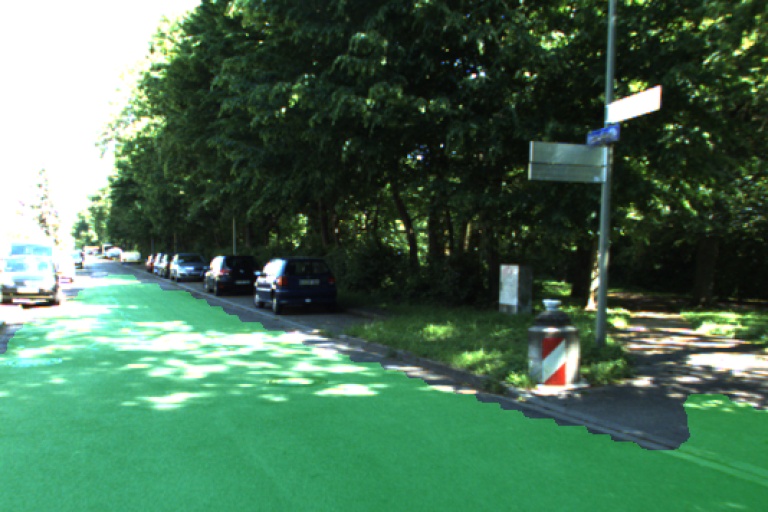} 
          & \includegraphics[width=0.18\linewidth]{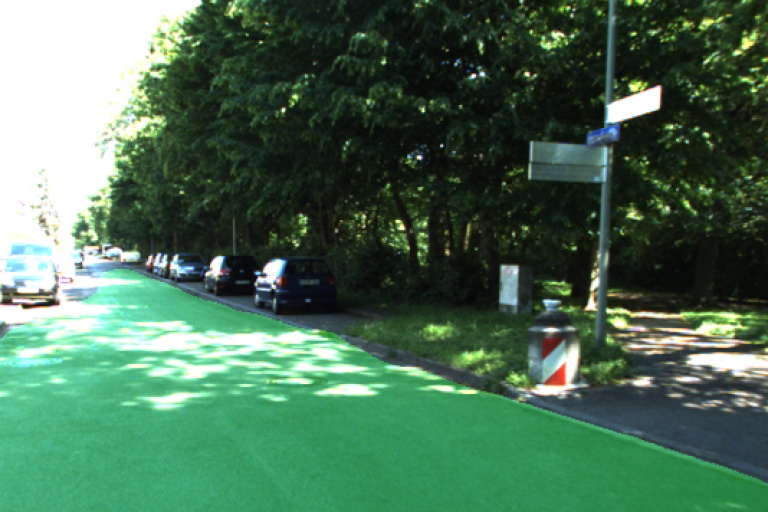} 
          & \includegraphics[width=0.18\linewidth]{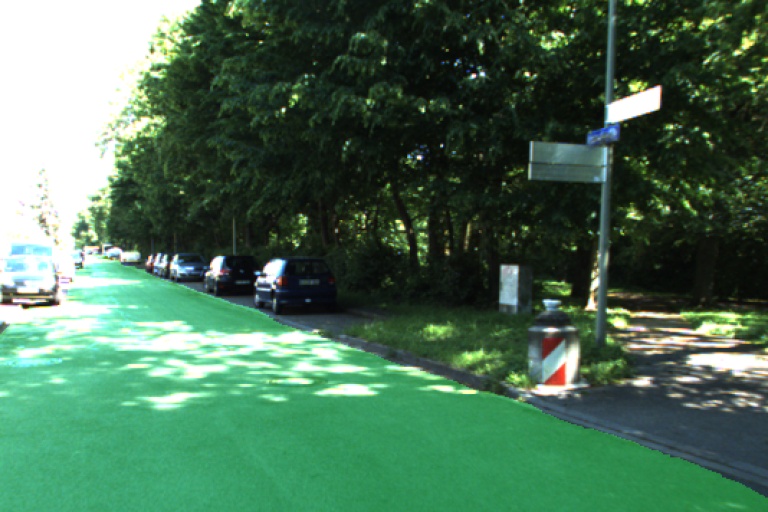} 
         \\
       \includegraphics[width=0.18\linewidth]{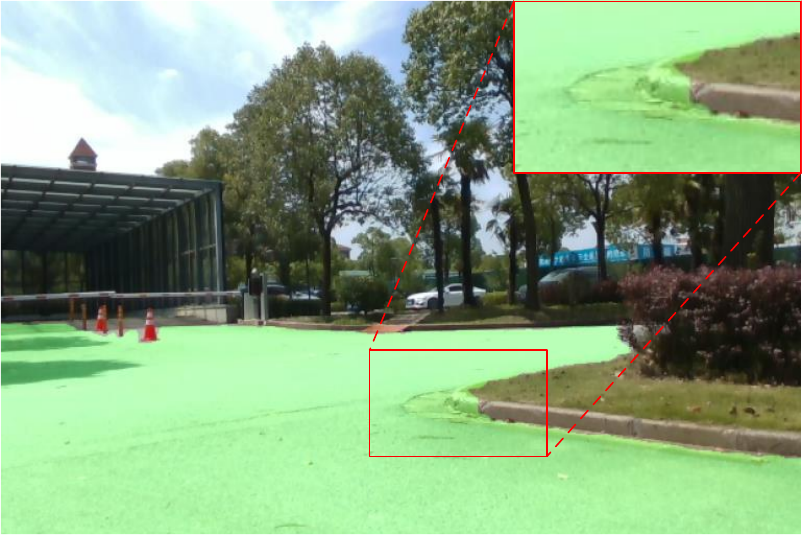} 
         & \includegraphics[width=0.18\linewidth]{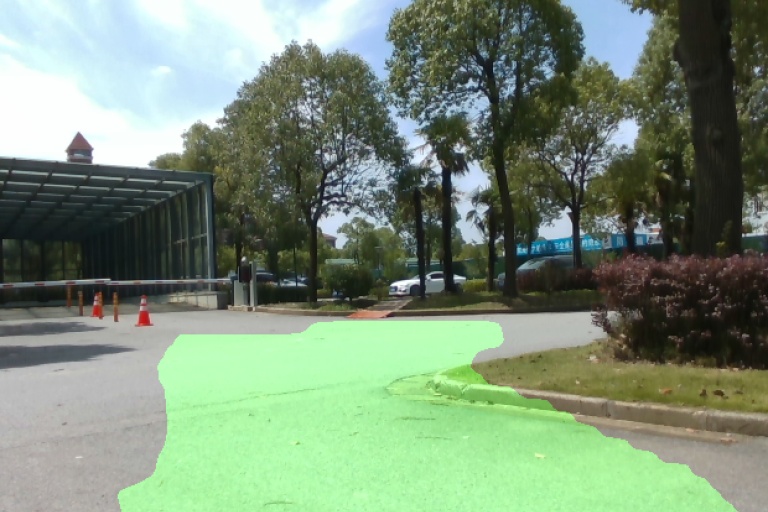} 
         & \includegraphics[width=0.18\linewidth]{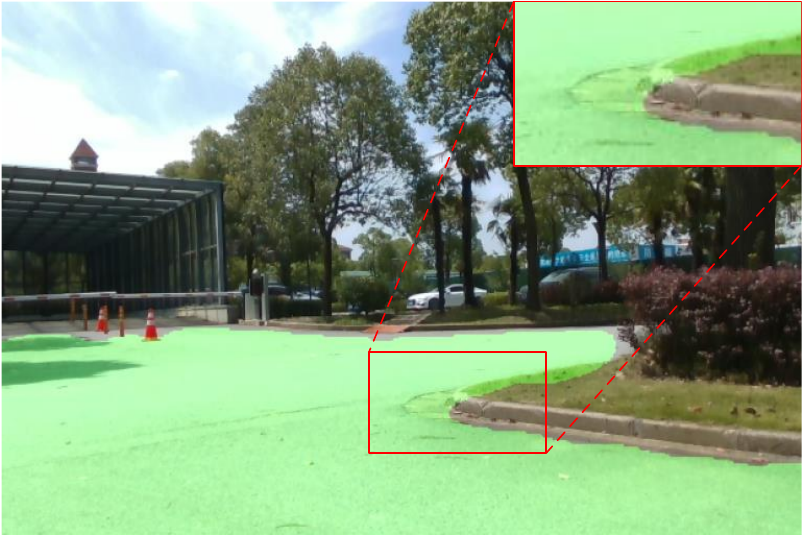} 
          & \includegraphics[width=0.18\linewidth]{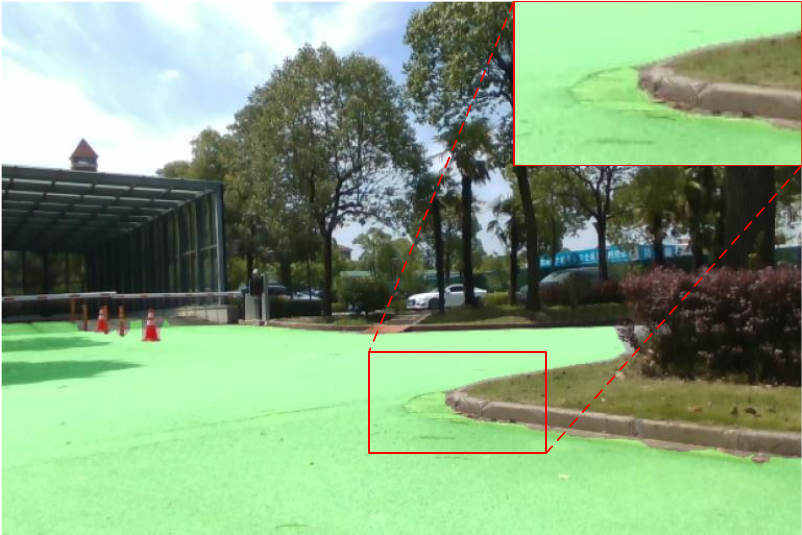} 
        & \includegraphics[width=0.18\linewidth]{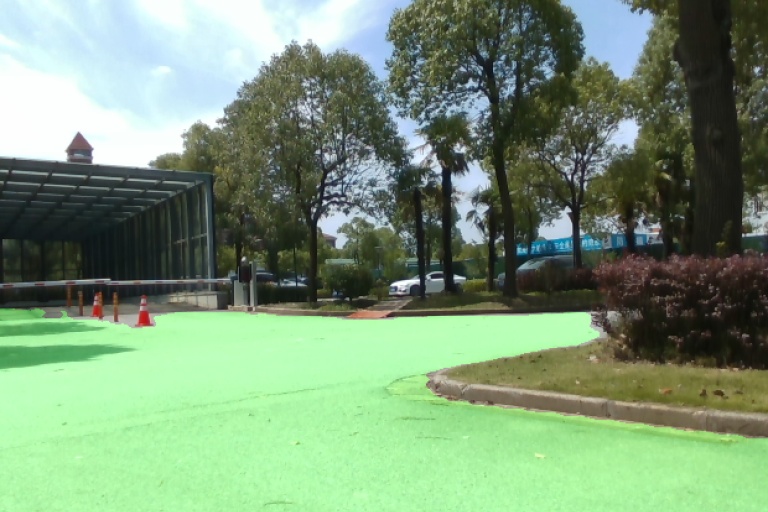} 
        \\
         \small{(a) EVAA} & \small{(b) FtFoot} & \small{(c) V-STRONG}  & \small{(d) Ours} & \small{(e) Ground Truth}
	\end{tabular}
	\caption{Qualitative results of different methods for traversability estimation. The first row corresponds to RELLIS-3D, the second to KITTI-360, and the third to our Campus dataset. Green pixels indicate traversable areas.}
	\label{fig:results_estimation}
\vspace{-5pt}
\end{figure*}

The \Tref{tab:results_estimation} and \Fref{fig:results_estimation} present the quantitative and qualitative results of traversability estimation across various datasets.

EVAA demonstrates strong performance on Campus dataset, characterized by clear spatial separation and continuous region structures. However, on more complex datasets such as RELLIS-3D and KITTI-360, the IoU decreases by 6.1\% to 11.1\% compared to Campus dataset. This performance degradation is primarily attributed to the limitations of the pseudo-label generation mechanism, as well as the training strategy that applies only a weak supervisory signal to negative samples via a low-weighted loss term.

FtFoot’s reliance on sparse footprint annotations without dense label expansion limits its performance. In RELLIS-3D and Campus dataset, its IoU stagnates around 65\%, mainly due to subtle geometric differences between traversable and non-traversable regions (e.g., roads vs. sidewalks) that weaken normal-based convolutional guidance. This highlights the need for explicitly decoupled integration of geometric and semantic cues, as implemented in our method.

V-STRONG leverages the powerful ViT-H backbone to extract rich semantic features and derive one traversability vector, but its contrastive learning relies on footprints and SAM-based labels that introduce noise, limiting prediction accuracy. Consequently, V-STRONG underperforms EVAA on the Campus dataset, mainly due to the diverse traversable surfaces in our data that challenge the representation capacity of its high-dimensional traversability vector.


In contrast, our approach consistently achieves higher IoU scores, improving by 1.6\% to 3.5\% across datasets, thanks to high-quality self-annotated labels and an efficient multimodal architecture. We observe that our network’s predictions achieve higher IoU than the annotated labels, demonstrating its ability to effectively filter label noise and learn meaningful knowledge.

\begin{table}[h!]
\centering
\caption{Quantitative results of different methods for traversability estimation. Best results are in \textbf{bold}.}
\resizebox{\linewidth}{!}{%
\begin{tabular}{llccccccc}
\toprule
Dataset & Method      & IoU   & FPR   & FNR   & AP    & F1  & Pre   & Rec   \\
\midrule

\multirow{4}{*}{RELLIS-3D} 
       & EVAA        & 0.887 & 0.061 & 0.032 & 0.917 & 0.940 & 0.914 & \textbf{0.968} \\
       & FtFoot      & 0.819 & 0.055 & 0.114 & 0.923 & 0.900 & 0.915 & 0.886 \\
       & V-STRONG    & 0.917 & 0.039 & \textbf{0.029} & 0.919 & 0.957 & 0.943 & 0.971 \\
       & Ours        & \textbf{0.943} & \textbf{0.016} & 0.034 & \textbf{0.939} & \textbf{0.971} & \textbf{0.976} & 0.966 \\

\midrule

\multirow{4}{*}{KITTI-360} 
       & EVAA        & 0.840 & 0.092 & 0.079 & 0.836 & 0.913 & 0.905 & \textbf{0.921} \\
       & FtFoot      & 0.674 & 0.113 & 0.208 & 0.796 & 0.805 & 0.819 & 0.792 \\
       & V-STRONG    & 0.879 & 0.063 & \textbf{0.063} & 0.896 & 0.936 & 0.934 & 0.937 \\
       & Ours        & \textbf{0.910} & \textbf{0.019} & 0.063 &  \textbf{0.912} & \textbf{0.953} & \textbf{0.969} & 0.937 \\

\midrule

\multirow{4}{*}{Campus} 
       & EVAA        & 0.945 & 0.018 & \textbf{0.013} & 0.905 & 0.972 & 0.957 & \textbf{0.987} \\
       & FtFoot      & 0.632 & 0.036 & 0.313 & 0.713 & 0.774 & 0.887 & 0.687 \\
       & V-STRONG    & 0.909 & 0.033 & 0.017 & 0.866 & 0.952 & 0.923 & 0.982 \\
       & Ours        & \textbf{0.960} & \textbf{0.010} & 0.018 & \textbf{0.915} & \textbf{0.980} & \textbf{0.977} & 0.982 \\

\bottomrule
\end{tabular}
}
\label{tab:results_estimation}
\vspace{-10pt}
\end{table}

\subsection{Real-World Deployment}\label{real-world}

To assess real-world applicability, we implement our approach on a wheelchair robot using a laptop with an i5-12400F CPU and an NVIDIA GeForce RTX 4060 GPU. The proposed traversability estimation network is integrated into the elevation mapping framework~\cite{miki2022elevation}, in combination with the geometric traversability layer. This framework generates a 2.5D elevation map that indicates the traversability status for each grid cell. With an inference time of approximately 20 ms, the network supports real-time sensor processing at 10 Hz without delay.

\begin{figure}[htbp!]
    \centering
    \includegraphics[width=\linewidth]{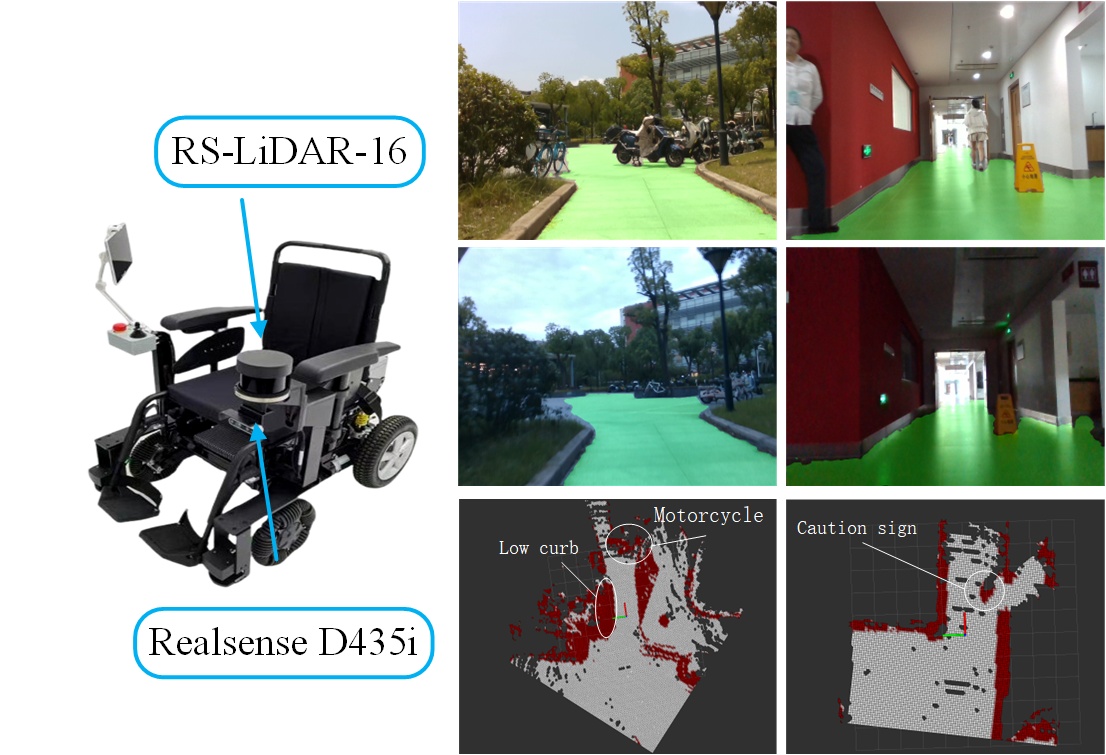}
    \caption{Real-world experiments. The first two rows present predictions across diverse environments (outdoor/indoor and varying illumination), and the final row shows the corresponding elevation mapping results in a top-down view for the scenes in the first row, where red indicates non-traversable regions.}
    \label{fig:real_experiments}
\vspace{-2mm}
\end{figure}

We evaluate our method in both outdoor and indoor environments, where it effectively handles irregular obstacles, dynamic pedestrians, low curbs, and varying lighting conditions, as is show in \Fref{fig:real_experiments}. This robustness is attributed to the LiDAR stream and diverse training data. Additional results are provided in the supplementary video.


\subsubsection{Effect of Automatic Traversability Labeling Component}
We conduct ablation studies on the refined prompt and the retrieval strategy proposed in our automatic traversability labeling pipeline. The result is shown in \Tref{tab:ablation_generation}.

\begin{table}[htbp!]
\centering

\caption{Ablation study results on automatic traversability labeling component. RP represents refined prompt, and RS represents retrieval strategy. Best results are in \textbf{bold}.}
\resizebox{0.9\linewidth}{!}{%
\label{tab:ablation_generation}
\begin{tabular}{llcccc}
\toprule
Dataset & Setting      & IoU  & F1  & Pre   & Rec \\
\midrule
\multirow{3}{*}{RELLIS-3D}
        & Ours              & \textbf{0.879}   & \textbf{0.936} & \textbf{0.927} & 0.944    \\
         & w/o RP                           & 0.838   & 0.912  & 0.913          & 0.910             \\
        & w/o RS                         & 0.873  & 0.932   & 0.900          & \textbf{0.968}        \\
       
\midrule
\multirow{3}{*}{KITTI-360}
        & Ours                & \textbf{0.883} & \textbf{0.938} & \textbf{0.964} & 0.914  \\
         & w/o RP                                 & 0.866 & 0.928    & 0.957          & 0.901         \\
        & w/o RS                 & 0.855  & 0.922    & 0.921          & \textbf{0.923}             \\
       
\bottomrule
\end{tabular}
}
\end{table}
Specifically, removing the refined prompt results in an IoU decrease of 1.9\% to 4.6\%, due to insufficient coverage of traversable areas and the lack of contrasting non-traversable examples necessary for SAM to fully understand the image.

Meanwhile, removing the retrieval strategy mainly causes a precision drop of 2.9\% to 4.4\% by failing to exclude negative regions. Although our retrieval strategy slightly reduces recall, but overall F1-score shows improvement. The lost recall can be compensated during network training via the refined loss, as discussed in the next section.

\subsubsection{Effect of Multimodal Traversability Estimation Network Component}
We further analyze the impact of our multimodal traversability estimation network design and the proposed refined loss through systematic ablation. The result is shown in \Tref{tab:ablation_estimation}.

\begin{table}[htbp!]
\centering
\caption{Ablation study results on multimodal traversability estimation network component on RELLIS-3D. Best results are in \textbf{bold}.}
\label{tab:ablation_estimation}
\resizebox{\linewidth}{!}{%
\begin{tabular}{lccccccc}
\toprule
 Setting            & IoU   & FPR   & FNR   & AP    & F1  & Pre   & Rec \\
\midrule
        Ours  & \textbf{0.943} & \textbf{0.015} & \textbf{0.034} & \textbf{0.939} & \textbf{0.970} & \textbf{0.975} & \textbf{0.965} \\
        w/o LiDAR branch & 0.919 & 0.022 & 0.049 & 0.919 & 0.958 & 0.966 & 0.950 \\
        w/o Refined Loss & 0.923 & 0.016 & 0.054 & 0.926 & 0.960 & 0.974 & 0.945\\
\bottomrule
\end{tabular}
}
\end{table}

When the LiDAR branch is removed, the model relies solely on RGB images, which lack geometric context. As a result, the precision is the lowest among all variants. Incorporating the LiDAR branch introduces valuable geometric cues, leading to improvements in both precision and overall metrics.

The refined loss compensates for regions that appear flat and traversable in LiDAR but are under-annotated in conservative SAM labels. Incorporating this loss boosts IoU by 2.1\% and lead to overall improved performance.

\section{Conclusions}

This paper presents a novel multimodal framework for self-supervised traversability labeling and estimation. In the traversability labeling
phase, visual foundation models and geometric priors are integrated to generate diverse positive and negative prompts for SAM, with a negative points guided retrieval strategy used to select the final mask. In the traversability estimation phase, the dual-stream network is trained using the refined loss that combines dense SAM labels and sparse LiDAR labels. Extensive experiments across multiple datasets and real-world platforms demonstrate that the proposed method significantly outperforms existing approaches in both automatic labeling quality and traversability estimation accuracy.

Future work includes exploring adaptive sampling strategies to address variations in point cloud density and image resolution. Moreover, more advanced multimodal fusion modules may be developed to tackle challenges arising from extreme weather conditions and sensor noise. Semi-supervised incremental learning will be investigated to further improve generalization. Additionally, advancements in foundation models may enable extending our approach from the 2D image domain to 3D spatial representations.

\addtolength{\textheight}{0cm}   

\bibliographystyle{IEEEtran}
\bibliography{mybib.bib}

\end{document}